\documentclass[twoside]{article}
\usepackage[accepted]{aistats2024}

\usepackage[round]{natbib}
\usepackage{graphicx}
\usepackage{algorithm}
\usepackage{algorithmic}
\usepackage{multirow}
\usepackage{url}

\begin{document}

%

%
\runningtitle{A White-Box FP Adversarial Attack on Contrastive Loss Based Signature Verification Models}
\runningauthor{Guo, Li, Qian, Arandjelovi\'c, Fang}
\twocolumn[

\aistatstitle{A White-Box False Positive Adversarial Attack Method on Contrastive Loss Based Offline Handwritten Signature Verification Models}


\aistatsauthor{Zhongliang Guo$^1$ \And Weiye Li \And Yifei Qian \And Ognjen Arandjelovi\'c$^2$ \And Lei Fang}



\aistatsaddress{University of St Andrews, St Andrews, United Kingdom; $^{1,2}$\{zg34,oa7\}@st-andrews.ac.uk}

]

\begin{abstract}
In this paper, we tackle the challenge of white-box false positive adversarial attacks on contrastive loss based offline handwritten signature verification models.
We propose a novel attack method that treats the attack as a style transfer between closely related but distinct writing styles.
To guide the generation of deceptive images, we introduce two new loss functions that enhance the attack success rate by perturbing the Euclidean distance between the embedding vectors of the original and synthesized samples, while ensuring minimal perturbations by reducing the difference between the generated image and the original image.
Our method demonstrates state-of-the-art performance in white-box attacks on contrastive loss based offline handwritten signature verification models, as evidenced by our experiments.
The key contributions of this paper include a novel false positive attack method, two new loss functions, effective style transfer in handwriting styles, and superior performance in white-box false positive attacks compared to other white-box attack methods.
\end{abstract}

\section{INTRODUCTION}

\begin{figure*}[tb]
    \centering
    \includegraphics[width=0.9\textwidth]{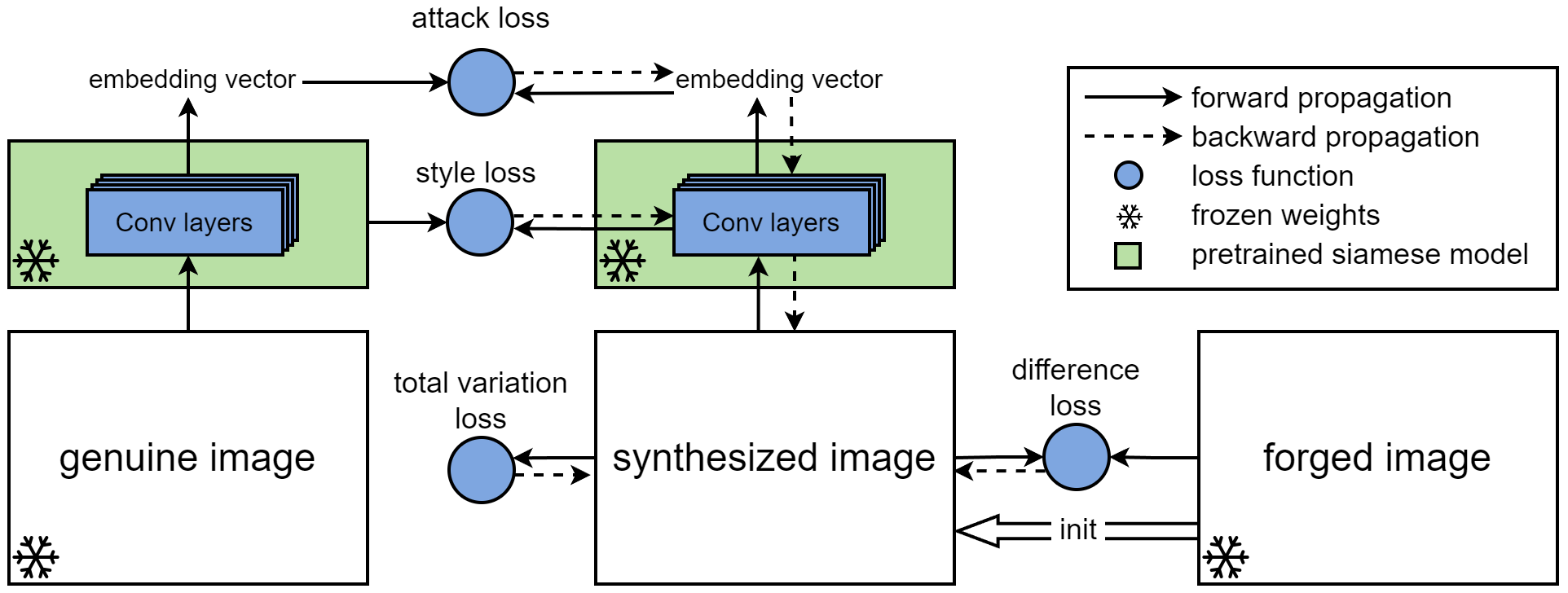}
    \caption{Overview of our approach to executing an attack on a contrastive loss based siamese neural network.
    The network architecture involves the processing of two inputs: a genuine and a synthesized image.
    The latter is initialized using content derived from a forged image, and uniquely, remains the sole trainable element within this otherwise frozen network setup.
    Upon processing, the network generates a pair of embedding vectors; these are integral in computing the attack loss.
    Simultaneously, style loss is calculated within designated convolutional layers distributed among the twin branches of the network.
    A difference loss emerges from the comparison between the synthesized and the forged images.
    Moreover, the synthesized image alone serves as the basis for the calculation of the total variation loss.
    Crucially, all computed losses are fed back into the synthesized image}
    \label{fig:arch}
\end{figure*}

In the ever-evolving digital landscape, adversarial attacks have become a serious concern.
An adversarial attack, which involves presenting manipulated input to machine learning models to induce incorrect outputs, has posed a threat to most models across the field
\citep{szegedy2013intriguing,liu2022mutual,liu2022segment,li2022adversarial,zhao2023prompt,lau2023interpolated,yu2023improving,guo2024artwork}.
Such attacks can be broadly classified into two categories: targeted and untargeted attacks.
In targeted attacks, the adversary tries to manipulate the system to produce a specific outcome.
In contrast, untargeted attacks aim to return an incorrect output, without any particular outcome in mind.
Additionally, depending on the adversary's knowledge about the system, attacks can be categorized as white-box attacks (where the attacker has full knowledge about the model) and black-box attacks (where the attacker has no knowledge about the model) \citep{papernot2016transferability,akhtar2018threat,kurakin2018adversarial}.
Adversarial attacks have become particularly important in the context of personal authentication, and in particular offline handwritten signature verification (hereafter referred to as signature verification), given the recent shift towards the use of biometric information.
Based on their output, we can distinguish between existing signature verification approaches based on classification models and contrastive learning models.
Classification models aim to classify signatures as genuine or forged based on extracted features \citep{yilmaz2015offline,hafemann2016writer,ren20232c2s}.
On the other hand, contrastive learning models aim to learn a representation where signatures from the same class are closer in the latent space~\citep{guo2023siamese}, while those from different classes are far apart \citep{rantzsch2016signature,dey2017signet}.

Notably, among these models, SigNet \citep{dey2017signet}, which employs contrastive learning, has consistently achieved state-of-the-art results across almost all public datasets.
Given its prominence and widespread use, we adopted it as our target model for adversarial attacks in the present work.

The extensive utilization of these systems has consequently increased their susceptibility to a variety of attack forms. Intriguingly, given the binary output of these systems -- authentic or forged -- the nature of attacks in this domain blurs the lines between targeted and untargeted adversarial attacks. Regardless of the attack strategy, the objective remains the same: to get a wrong decision from targeted systems. In the context of this work, the types of attack of interest are True Negative (TN) and False Positive (FP) attacks.

TN attack is aimed at authentic signatures.
The goal is to manipulate a genuine signature such that it is falsely rejected by the model as a forgery. Conversely, the FP attack targets forged signatures, whose aim is to subtly modify a forged signature to the point where it is falsely accepted by the model as being genuine.

The existing attack methods, such as Fast Gradient Sign Method (FGSM), Iterative Gradient Sign Method (IGS), and Momentum Iterative Method (MIM), were primarily designed for classification tasks on images with richer backgrounds, making them less effective on signature images which typically have clean backgrounds \citep{goodfellow2014explaining,kurakin2018adversarial,dong2018boosting}.
This limitation is further highlighted by the evident imbalance between the success rates of TN and FP attacks \citep{hafemann2019characterizing}. The clean background of signature images makes any introduced perturbation stand out, diminishing the successes of FP attacks. This observation has significantly driven our research focus towards FP attacks.

Rather than promoting these attacks, our study seeks to understand their underlying mechanics and their impact on contrastive loss based signature verification systems. The goal is to identify weaknesses in current authentication methods and further develop countermeasures to enhance system robustness and security.

The signature verification systems that we target in this study are predominantly trained with contrastive loss \citep{hadsell2006dimensionality}, a strategy that has proven highly effective in learning rich representations for signature verification.
To our knowledge, there are currently no attack methods specifically designed for contrastive loss based models.
Although various attack methods intended for classification models can be adapted for these systems, their effectiveness is greatly reduced in FP attacks, as our experimental results will shortly demonstrate.
This highlights the need for an effective FP attack method on contrastive loss based models, which will contribute to the development of more robust and reliable authentication systems.

To this end, in this paper we consider style transfer as a potentially beneficial tool. Style transfer, the task of applying the style of one image to the content of another, has seen impressive advancements over the years. Pioneering work by \citet{gatys2015neural} used Convolutional Neural Networks (CNN) to separate and recombine content and style of arbitrary images, resulting in artistically appealing results \citep{gatys2015neural,johnson2016perceptual,zhu2017unpaired}.
Inspired by their work, we propose to think of signature verification as a comparison of writing styles with identical content.
By harnessing the power of style transfer, we aim to bridge the gap between different writing styles through the perturbations generated by our attack.
By aligning the visual attributes of manipulated signatures with those of genuine signatures, we strive to deceive the model by blurring the distinction between authentic and forged writing styles.
We believe that this approach can significantly contribute to enhancing the success rate of FP attacks.

This research serves as a stepping stone towards bolstering the security of biometric systems.
Our focus on FP attacks and the application of style transfer aim to challenge existing vulnerabilities, paving the way for more comprehensive security measures in biometric systems, specifically those using contrastive learning.

The main contributions of the present are:
\begin{itemize}
    \item We propose a novel FP attack method to address the inability of existing attack methods to contrastive loss based signature verification systems.
    \item We propose two novel loss functions which guide the generation of deceptive images by minimizing the distance between the embedding vectors of the genuine and synthesized images, meanwhile ensuring minimal perturbations.
    \item We demonstrate the effectiveness of style transfer in terms of handwriting style.
    This is achieved by the success of embedding the writing-style texture of the target signature into the perturbation.
    \item We present state-of-the-art performance in executing white-box FP attacks, greatly outperforming the existing white-box attack methods.
\end{itemize}

\section{RELATED WORK}
The field of adversarial attacks and its defense on signature verification systems, though of significant importance, remains under-explored. So far, only a small amount of work has been done.

\citet{yu2016two} focused on optimizing decision-making in signature verification systems in the presence of spoofing attacks.
They explored different optimization strategies to enhance the robustness of classification models used in signature verification.
Their research provided valuable insights into potential approaches for improving the reliability of signature verification systems in real-world scenarios.
\citet{hafemann2019characterizing} conducted a study characterizing and evaluating adversarial examples for signature verification, with a specific focus on the impact of different classification heads .
They investigated the vulnerabilities of signature verification systems to adversarial attacks and found an imbalance in the success rates of True Negative (TN) and False Positive (FP) attacks in classification models.
This highlights the importance of addressing the robustness and balance of FP attacks in signature verification, which is one of the motivations for our research.
\citet{li2021black} proposed the first black-box attack method against signature verification systems.
They demonstrated the vulnerability of signature verification models to black-box attacks and introduced a novel approach to generate adversarial perturbations that are restricted to specific regions of the signature image.
Notably, their targeted model is also a classification model rather than a contrastive learning model.
\citet{bird2023writer} investigated the vulnerabilities of signature verification systems to robotic.
Their research highlighted the potential risks associated with robotic writers in signature verification and revealed the limitations of using GANs for generating signatures with similar distributions.
Their findings emphasize the need for robust defenses in signature verification systems to mitigate these threats.

A common thread in these studies is their focus on classification models, revealing a significant gap in research concerning adversarial attacks on other types of models in signature verification systems.
This calls for further research to address this void and diversify defenses against adversarial attacks.

\section{METHODS}

In this section, we introduce our novel approach to executing an attack on a contrastive loss based Siamese neural network.
As outlined in Figure \ref{fig:arch}, the method incorporates four distinct loss functions in the generation of misleading synthesized images from a forged image.
These loss functions guide the adversarial training process by encouraging the feature representations of the synthesized images to move closer to those of genuine images.
Our attack is white-box due to its superior performance for adversarial research and the ease of accessing network information through hardware-level attacks~\citep{wei2020leaky,zhang2021stealing}.

Firstly, we will define the problem and the specific type of attacks our method aims to carry out.
Then, we will dive into a detailed explanation of the loss functions and optimization involved in our approach.

\subsection{Problem Definition}

Our method pivots from the conventional approach of attacking classification models to target those based on contrastive loss.
Our main focus is on signature verification models.
The operation process of the model \citep{dey2017signet} we aim to attack unfolds as follows.

The model applies a threshold $\tau$ on the distance measure $D(x_i, x_j)$, to segregate signature pairs $(i, j)$ into classes of similarity or dissimilarity.
Pairs possessing identical identities are labeled $P_{similar}$, while pairs with different identities are classified as $P_{dissimilar}$. In this framework, the collection of all true positives (TP) at the threshold $\tau$ is expressed as:
\begin{align}
    \textnormal{TP}(d) = \{(i, j) | (i, j) \in P_{similar} \textnormal{ and } D(x_i, x_j) \leq \tau\} \notag.
\end{align}
Similarly, the set of all true negatives (TN) at distance $\tau$ is defined as:
\begin{align}
\text{TN}(\tau) = \{(i, j) | (i, j) \in P_{dissimilar} \text{ and } D(x_i, x_j) > \tau\} \notag.
\end{align}
Subsequently, the true positive rate $\text{TPR}(\tau)$ and the true negative rate $\textnormal{TNR}(\tau)$ for a specific signature distance $\tau$ are computed as follows:
\begin{align}
\text{TPR}(\tau) = \frac{|\text{TP}(\tau)|}{|P_{similar}|},\ \ \text{TNR}(\tau) = \frac{|\text{TN}(\tau)|}{|P_{dissimilar}|} \notag,
\end{align}
where $P_{similar}$ represents the total count of similar signature pairs. The final accuracy is determined as:
\begin{align}
  \text{Accuracy} = \max_{\tau \in D} \left(\frac{1}{2} \cdot (\text{TPR}(\tau) + \text{TNR}(\tau))\right) \notag.
\end{align}
This is the maximal accuracy achievable by adjusting $\tau$ from the smallest to the largest distance value in $D$, with an incremental step.

In our method, we strategically choose to focus on FP attacks due to the asymmetry in difficulty between FP and TN attacks.
Our goal is to craft an algorithm that can generate deceptive forgeries which would lead to a decrease in the model's verification accuracy.

Let us consider a FP attack method towards the above contrastive loss based signature verification model.
Define a set $\mathcal{S}$, where each element is a pair of images $(\mathbf{Y_i}, \mathbf{Y_j})$ such that:
\begin{align*}
\mathcal{S} = \{(\mathbf{Y_i}, \mathbf{Y_j}) \mid (\mathbf{X_i}, \mathbf{X_j}) = F(\mathbf{Y_i}, \mathbf{Y_j}), D(\mathbf{X_i}, \mathbf{X_j}) > \tau\} \notag.
\end{align*}

In $\mathcal{S}$, $(\mathbf{X_i}, \mathbf{X_j})$ are d-dimensional vectors computed through a function $F$ (representing the signature verification model), and $D$ is the distance measure function in the embedding space.
The decision threshold, $\tau$, is a constant.

Our objective is to find a transformation function $G$, which has $\mathbf{Y_j^\prime}=G(\mathbf{Y_j})$, such that a new set $\mathcal{S'}$ is defined as follows:
\begin{align*}
\mathcal{S'} = \{(\mathbf{Y_i}, \mathbf{Y_j^\prime}) \mid (\mathbf{X_i}, \mathbf{X'_j}) = F(\mathbf{Y_i}, \mathbf{Y_j^\prime}), D(\mathbf{X_i}, X'_j) \leq \tau\} \notag,
\end{align*}
where $\mathbf{X'_j}$ represents the new embedding vector obtained from the transformed image $G(\mathbf{Y_j})$. The function $G$ modifies the original image $\mathbf{Y_j}$ to create a new image, but does not change $\mathbf{Y_i}$ or the system parameters. The goal is to ensure that for all pairs in $\mathcal{S'}$, the distance between the embedding vectors is less than $\tau$.

\subsection{Attack Loss}

The proposed attack loss function is designed to guide the generation of deceptive images through adversarial training.
It operates by encouraging the reduction of the Euclidean distance between the vector representations of the synthesized image, denoted by $\mathbf{X_2}$, and the genuine image, denoted by $\mathbf{X_1}$.
Both $\mathbf{X_1}$ and $\mathbf{X_2}$ are calculated from the genuine image and the synthesized image through the pre-trained model.
\begin{equation}\label{equ:attackloss}
\begin{aligned}
L(\mathbf{X_{1}},\mathbf{X_{2}},\tau) &=\\ &I(\left\|\mathbf{X_{1}}-\mathbf{X_{2}}\right\|_2>\tau) \frac{\left\|\mathbf{X_{1}}-\mathbf{X_{2}}\right\|_2}{\tau} .
\end{aligned}
\end{equation}

The function employs a binary indicator $I$ that becomes active (i.e., equals 1) when the Euclidean distance between $\mathbf{X_1}$ and $\mathbf{X_2}$ is greater than a given threshold $\tau$.
In such cases, the loss is computed as the ratio of the Euclidean distance to the threshold $\tau$, effectively representing the degree to which the current distance exceeds the acceptable threshold.
The loss function, $L$, is defined as in \eqref{equ:attackloss}.

By minimizing this loss over a number of training iterations (epochs), the adversarial training process gradually adjusts the synthesized image to decrease its Euclidean distance to the genuine image.
The aim is that, after sufficient training, the synthesized image is close enough to the target in the vector space to deceive the model, while still maintaining a discernible distance from the genuine in the image space.
Thus, this loss function is key to our method for generating adversarial samples.
It allows us to create images that, after sufficient training iterations, are close enough to the target in the vector space to deceive the model, while still preserving a reasonable level of variation from their initial state, avoiding directly ``copy'' the genuine image.

\subsection{Difference Loss}

The proposed difference Loss function, denoted as L, aims to minimize the discrepancies between the synthesized image, represented by $\mathbf{X_2}$, and its initial state, represented by $\mathbf{X_1}$.
It accomplishes this objective by calculating the absolute differences between corresponding elements of the two vectors and summing them up:
\begin{equation}\label{equ:diffloss}
\textnormal{L}\left ( \mathbf{X_1} ,\mathbf{X_2} \right ) = \sum_{i=1}^{N} \left | \mathbf{X_{1_i}} - \mathbf{X_{2_i}}\right |.
\end{equation}

The primary purpose of the Difference Loss is to ensure that the synthesized image retains a high degree of similarity to its initial state.
By minimizing this loss, we strive to generate adversarial samples that introduce minimal perturbations, as our objective is to exploit the network's vulnerability to the smallest possible perturbation that can lead to incorrect decisions.
Consequently, the synthesized images exhibit subtle differences from the genuine ones, enhancing the stealthiness of the attack.

Minimizing the discrepancies between the synthesized image and its initial state using the Difference Loss enables us to craft adversarial samples with minimal visual alterations, thereby maximizing the effectiveness of our attack strategy.

\subsection{Style Transfer}

Inspired by the concept of neural style transfer \citep{gatys2015neural}, we incorporate the concept into our attack strategy to address the distinctive challenges posed by signature verification.
We treat the verification problem as comparing two different but similar writing styles corresponding to a consistent content.
Thus, a genuine signature and its corresponding forged signature can be viewed as two images that share identical content but slightly different styles.
With this understanding, we introduce the style loss and the total variation loss components into our attack architecture to generate the perturbations necessary to bridge this stylistic gap.

\subsubsection{Style Loss}

The style loss is computed as the mean squared error between the Gram matrices of the feature representations of the genuine and synthesized images.
Gram matrices serve as a representation of style, encapsulating the distribution of feature correlations at each layer of the network.
By minimizing the difference in these matrices, we aim to align the styles of the genuine and synthesized images.
This allows us to create a synthesized image that retains the content of the genuine image while adopting the style of the target image. The loss is:
\begin{equation}\label{equ:styleloss}
L_{style} (\mathbf{F_{1}},\mathbf{F_{2}}) = \frac{1}{4N^2M^2} \sum_{i=1}^{N} \sum_{j=1}^{M} (\mathbf{G}_{ij}^{\mathbf{F_{1}}}-\mathbf{G}_{ij}^{\mathbf{F_{2}}})^2,
\end{equation}
where $\mathbf{F_{1}}$ and $\mathbf{F_{2}}$ are the feature representations of the genuine and synthesized images, respectively.
$N$ is the number of feature maps, $M$ is the height times the width of the feature map, and $\mathbf{G}_{ij}$ represents the elements of the Gram matrix, which are computed by taking an outer product of the vectorized feature maps and then averaging over all positions.

Minimizing this loss function encourages the synthesized image to adopt a style that closely mimics that of the genuine image.
This results in a synthesized image that bears a strong stylistic resemblance to the genuine, thereby confusing the model and increasing the success rate of the attack.
This demonstrates the practical value of style transfer in revealing potential vulnerabilities in machine learning models, extending its application beyond the realm of artistic image generation.

Through the application of style loss, our method successfully bridges the stylistic gap between the genuine and synthesized images, making them more similar to each other in style.
This strategy effectively tricks the model into making wrong decisions by generating less obvious subtle perturbations, thereby increasing the success rate of the attack.
By employing style transfer in the context of adversarial attacks, we extend its application beyond the realm of artistic image generation, demonstrating its practical value in exposing potential vulnerabilities in machine learning models.

\subsubsection{Total Variation Loss}
Total Variation Loss (TV Loss) plays a crucial role in style transfer tasks.
TV loss is typically used as a regularization term in the loss function.
By penalizing large variations between neighboring pixel values, it helps reduce high frequency noise and produce more visually appealing results.

Let's denote the synthesized image as $Y$.
Then, the Total Variation Loss can be written as follows:
\begin{equation}\label{equ:tvloss}
\begin{aligned}
L_{TV}(\mathbf{Y}) = \sum_{i,j} ( (\mathbf{Y}_{i,j+1} - \mathbf{Y}_{i,j})^2 +\\
(\mathbf{Y}_{i+1,j} - \mathbf{Y}_{i,j})^2 ),
\end{aligned}
\end{equation}
where $Y_{i,j}$ represents the pixel value at location $(i,j)$ in the synthesized image $Y$.
This loss function computes the sum of the squares of differences between neighboring pixel values in both horizontal and vertical directions.
By minimizing this loss, we encourage smooth transitions between neighboring pixels, thereby reducing high-frequency noise and producing more visually coherent synthesized images.

Adding the TV Loss to our objective not only helps improve the visual quality of the synthesized images, but also imposes an additional constraint on the solution space, making it more difficult for the model to distinguish between the genuine and synthesized images, thus increasing the attack success rate.

\subsection{Total Loss}

To strike a balance among all these considerations and drive the model towards the generation of effective deceptive images, we introduce a composite loss function --- the Total Loss.
The Total Loss function is a weighted sum of the Attack Loss, Difference Loss, Style Loss, and Total Variation Loss, combining their individual strengths to guide the adversarial training. It can be formulated as follows:
\begin{equation}\label{equ:totalloss}
L_{total} = \alpha L_{attack} + \beta L_{di\!f\!ference} + \gamma L_{style} + \delta L_{TV},
\end{equation}
where $L_{attack}$, $L_{di\!f\!ference}$, $L_{style}$, and $L_{TV}$ represent the Attack Loss, Difference Loss, Style Loss, and TV Loss respectively.
The terms $\alpha$, $\beta$, $\gamma$, and $\delta$ are weighting factors that control the contribution of each individual loss to the total loss.

By judiciously selecting these weights, we can influence the adversarial training process to place varying emphasis on matching the vector representations (Attack Loss), maintaining similarity to the initial image state (Difference Loss), aligning the style of the images (Style Loss), and promoting visual smoothness (Total Variation Loss).

By minimizing the Total Loss, we are able to generate adversarial images that preserve a high level of stealth and visual consistency while still effectively confusing the signature verification model.
This indicates that our method is capable of crafting precise, targeted adversarial attacks aimed at a specific objective.
Our approach underscores the importance of considering all relevant factors when formulating an attack strategy, including retaining the genuine style of the synthesized image, minimizing noticeable changes to the image, and getting as close as possible to the target in the image space, not just in the feature representation.
This comprehensive attack strategy allows our method to deceive the model more effectively, thereby increasing the attack success rate.

\subsection{Optimization}

\begin{algorithm}[tb]
\caption{Pseudocode of the proposed method.}
\label{alg:attack}
\textbf{Input}: genuine image and forged image $I_{g}, I_{f}$\\
\textbf{Parameter}: network $F$ parameters $W$; weights for total loss $\alpha,\beta,\gamma,\delta$; threshold for attack loss $\tau$; selected layers for $L_{style}$ $\zeta$ \\
\textbf{Output}: synthesized image $I_{gen}$

\begin{algorithmic}[1] 
\STATE Initialize: $I_{gen} \leftarrow I_f$
\STATE Initialize: $L_{attack} \leftarrow \tau$
\STATE Initialize: $L_{style} \leftarrow \zeta$
\STATE Initialize: $L_{total}\leftarrow (\alpha,\beta,\gamma,\delta)$
\STATE $f\!eatures_g,embed_g = F_{W}(I_g)$
\FOR{each epoch $e$}
    \STATE $f\!eatures_f,embed_f = F_{W}(I_f)$
    \STATE $l_{attack} = L_{attack}(embed_g,embed_f)$
    \STATE $l_{tv} = L_{TV}(I_{gen})$
    \STATE $l_{style} = L_{style}(f\!eatures_g,f\!eatures_f)$
    \STATE $l_{di\!f\!ference} = L_{di\!f\!ference}(I_{f},I_{gen})$
    \STATE $l_{total} = L_{total}(l_{attack}, l_{tv}, l_{style},l_{di\!f\!ference})$
    \STATE Update $I_{gen}$ via Adam optimization algorithm
\ENDFOR
\STATE \textbf{return} $I_{gen}$
\end{algorithmic}
\end{algorithm}

To optimize our composite Total Loss function, we employ the Adam optimization algorithm \citep{kingma2014adam}, which is widely used for training deep learning models due to its ability to handle sparse gradients on noisy problems.

Our method iteratively refines the synthesized image over multiple epochs.
In each epoch, we calculate the Total Loss and backpropagate it to update the image.
Note that only the synthesized image is trainable, while the weights of the Siamese neural network remain frozen.

After several training epochs, the synthesized image becomes a deceptive image that is visually similar to the genuine image yet is close enough in the vector space to the target image to deceive the model.
By systematically tuning the loss weights and the threshold, we can control the trade-off between the attack success rate and the perceptibility of the attack. The pseudocode in Algorithm~\ref{alg:attack} summarizes the proposed attack method.

This algorithm describes the overall process of adversarial attack on contrastive loss based models.
Starting with an genuine and a forged image, we iteratively optimize a synthesized image to minimize the Total Loss.
As a result, we generate a synthesized image that is both visually similar to the genuine image and likely to deceive the model.


\begin{table*}[tb]
\caption{Experimental settings and results. ``ST'' indicates the style transfer, ``FG'' indicates the foreground of the image, ``AT'' indicates the defense method, adversarial training. The VMI-FGSM is with $\mu=0.9,N=5$.}
\label{tab:eval}
\begin{center}
\begin{tabular}{lrrrr}
\hline
\multicolumn{1}{c}{}                     & \multicolumn{4}{c}{DATASETS}                                                 \\
\multicolumn{1}{c}{}                     & \multicolumn{2}{c}{CEDAR}  & \multicolumn{2}{c}{BHSig260-B}            \\
\multicolumn{1}{c}{\multirow{-3}{*}{ATTACK METHOD}} & TN ATTACK & FP ATTACK & TN ATTACK & FP ATTACK \\ \hline
PGD ($\epsilon=0.3,\alpha=0.1$) & 44.57\% & 0\% & 100\% & 0\% \\
PGD ($\epsilon=0.05,\alpha=0.01$) & 0.07\% & 0\% & 68.42\% & 0\% \\
PGD w/ ST ($\epsilon=0.3,\alpha=0.1$) & 98.77\% & 1.23\% & 99.97\% & 0.03\% \\
PGD w/ ST ($\epsilon=0.05,\alpha=0.01$) & 99.93\% & 0\% & 97.90\% & 2.01\% \\
PGD w/ FG ($\epsilon=0.3,\alpha=0.1$) & 100\% & 0\% & 20.34\% & 0\% \\
PGD w/ FG ($\epsilon=0.05,\alpha=0.01$) & 100\% & 0\% & 72.60\% & 0\% \\
VMI-FGSM ($\epsilon=0.3,\beta=3$) & 100\% & 0\% & 100\% & 0\% \\
VMI-FGSM ($\epsilon=0.05,\beta=3$) & 100\% & 0\% & 100\% & 0\% \\
VMI-FGSM ($\epsilon=0.3,\beta=1.5$) & 100\% & 0\% & 100\% & 0\% \\
VMI-FGSM ($\epsilon=0.05,\beta=1.5$) & 100\% & 0\% & 100\% & 0\% \\
C\&W ($c=0.1$)  &100\%  &0\%  &  96.20\%  &  0\%\\
C\&W ($c=0$)  &0\%  &0\%  &  57.98\%  &  0\%\\
MIM ($\epsilon=0.3$, $\mu=0.9$)    & 100\%   & 0\%              & 100\%                        & 0\%              \\
MIM ($\epsilon=0.05$, $\mu=0.9$)   & 0\%     & 0\%              & 86.81\%                      & 0\%              \\
IGS ($\epsilon=0.3$)             & 100\%   & 0\%              & 100\%                        & 0\%              \\
IGS ($\epsilon=0.05$)            & 20.51\% & 0\%              & 100\%                        & 0\%              \\
FGSM ($\epsilon=0.3$)                      & 100\%   & 0\%              & 98.84\%                      & 0\%              \\
FGSM ($\epsilon=0.05$)                     & 100\% & 0\%              & 81.44\%                      & 0\%              \\
\textbf{Ours}           & ---     & \textbf{99.71\%} & ---                          & \textbf{82.93\%} \\
Ours w/o ST & ---     & 92.32\%          & ---                          & 65.02\%          \\ 
Ours after AT & ---     & 93.70\%          & ---                          & 51.80\%          \\\hline
\end{tabular}
\end{center}
\end{table*}

\section{RESULTS}
In this section, we detail the experiments conducted to evaluate the performance of our proposed adversarial attack method, as well as the subsequent results.
We first present the datasets used in our experiments, followed by the pre-processing and model training procedures.
Then we discuss the specific settings for each attack method tested.
Finally, we analyze and evaluate the resulting performance metrics of each method.
These results provide a crucial understanding of the comparative effectiveness of our proposed adversarial attack strategy compared to existing methods.
\begin{figure*}[tb]
    \centering
    \includegraphics[width=\textwidth]{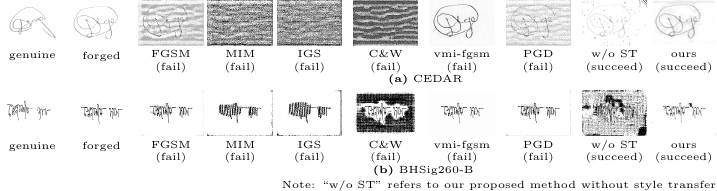}
    \caption{Visual results of several attack methods on two pairs of images.}
    \label{f:visRes}
\end{figure*}

\subsection{Dataset and Pre-trained Model}

We used two data sets in our experiments, the CEDAR\footnote{Available at \url{https://cedar.buffalo.edu/NIJ}} and BHSig260-B\footnote{Available at \url{https://goo.gl/9QfByd}}, each offering a broad spectrum of signature styles.
The CEDAR database features 1,320 genuine and 1,320 forged signatures, while the BHSig260-B, focusing on Bengali script signatures, offers signatures from 160 individual writers, with 24 authentic and 30 forged ones per writer.

Our data split follows the approach of our target model and is divided into trainset and testset. M out of K authors are randomly selected for training, while the remaining authors are reserved for testing. To ensure balance, we generate an equal number of genuine and forged pairs. This results in M × 276 training pairs and (K - M) × 276 testing pairs for each class.

The trainset is used to obtain the pre-trained model and its decision threshold $\tau$ is used in the attack, while the testset is used in developing the attack method.

In following the procedures of the target model, we utilize the trainset to acquire the pre-trained model and its associated decision threshold $\tau$ for each dataset.
The number of image pairs for FP attack and $\tau$ for each dataset is shown in Table~\ref{tab:attackpairs}.

\begin{table}[ht]
\caption{The decision threshold $\tau$ and the number of data pairs for FP Attack.}\label{tab:attackpairs}
\begin{center}
\begin{tabular}{lrr}
\hline
\multicolumn{1}{c}{DATASET} & \multicolumn{1}{c}{$\mathbf{\tau}$} & \multicolumn{1}{c}{FP PAIRS}\\\hline
CEDAR      & 0.0314           & 1380            \\
BHSig260-B & 0.03             & 9433            \\ \hline
\end{tabular}%
\end{center}
\end{table}

After the model training and threshold determination, we applied the model to the testset.
Only those signature pairs correctly decided by the model were selected for the following adversarial attacks.

\subsection{Experimental Setup}
We compare our proposed adversarial attack method\footnote{Implementation is available at \url{https://github.com/ZhongliangGuo/FP-attack}} against the conventional methods
FGSM~\citep{goodfellow2014explaining},
C\&W~\citep{carlini2017towards},
IGS~\citep{dong2018boosting},
MIM~\citep{kurakin2018adversarial},
PGD~\citep{madry2018towards},
and VMI-FGSM~\citep{wang2021enhancing}.
To evaluate the effectiveness of the style transfer component, we also incorporate it into PGD.
We rule out the negative effect of the background by restricting the perturbations solely on positions where have pixel bigger than 155/255.
To adapt to our problem, all attack methods use the Contrastive Loss to make them suitable for our target models. In the C\&W method, we replaced the parameter $\kappa$, meant for classification tasks, with a norm-based metric.

The experimental settings for each method, shown in Table~\ref{tab:eval}, were chosen based on their reported performance in literature and relevance to our work. We experimented with two sets of parameters. The first set was chosen to represent aggressive attack parameters, with high epsilon ($\epsilon=0.3$, for C\&W, it is $c=0$). The second set, with lower epsilon ($\epsilon=0.05$, for C\&W, it is $c=1$), represents a more conservative setting.

For our proposed method, three different experiments were conducted.
The first experiment leverages the full extent of our proposed method, incorporating the style transfer component.
The second experiment, however, omits the style transformation step.
The intention behind this approach is to determine the specific contribution of style transformation to the overall effectiveness of our attack method.
By comparing the results of the two experiments, we can distinctly observe the impact of style transformation on the success of our adversarial attack.
For the last experiment,
following the attack with style transfer, we further employed the generated adversarial samples in the adversarial training phase. This was done to assess the robustness of our attack method when faced with enhanced defensive measures. During adversarial training phase, the training data comprised a mix of normal data and adversarial samples at a ratio of 7:3.

It should be noted that, for all iterative methods, the number of epochs was fixed at 50.
This parameter was set based on preliminary experiments, aiming to balance the training time and the performance of attacks.

\subsection{Results and Discussion}

Upon analyzing the evaluation results displayed in Table \ref{tab:eval}, a clear distinction emerges between the performance of conventional adversarial attack methods and our proposed method when targeting signature verification models based on Contrastive Loss.
The conventional methods, originally designed for classification models with intricate backgrounds, achieve commendable success rates in TN attacks.

However, when it comes to signature verification, where images have clean backgrounds, their success rate plummets to 0\% for FP attacks.
Compared to images in other tasks, signature images lack many features, primarily focusing on the foreground with minimal background elements.
The feature set leads to heightened sensitivity to noise.
When noise is introduced, it can convert mismatched local features into seemingly matched ones, but also cause the model to misinterpret the overall image scale.

In stark contrast, our method treats the image itself as a trainable component.
Incorporating our proposed novel loss functions, manages to overcome this challenge.
Figure~\ref{f:visRes} demonstrates that compared with others, our method even without style transfer, does not change strokes. Such that maintains image scale consistency while introduced noise compensates for local feature mismatches, which is subject to the translation invariance of CNN. The introduce of style transfer further aligns the overall scale and feature positions, resulting in even better visual result.

This approach allows us to introduce subtle perturbations that correct non-matching local features without causing significant global changes.
Quantitatively, even in the absence of style transfer, our approach achieves commendable success rates in FP attacks -- 92.32\% for the CEDAR dataset and 65.02\% for the BHSig260-B dataset.
Furthermore, even when faced with models that have undergone adversarial training as a defensive measure, our method still demonstrates impressive success rates: 93.7\% on CEDAR and 51.8\% on BHSig260-B.

Further bolstering our method's effectiveness, the inclusion of style transfer leads to a significant increase in FP attack success rates.
With the synergistic combination of our loss functions and style transfer applied over 50 epochs, our method achieves an impressive 99.71\% success rate in FP attacks on the CEDAR dataset and a substantial 82.93\% success rate on BHSig260-B. Style transfer has also had a positive impact on traditional methods. We notice that integrating style transfer into PGD significantly enhances its TN attack.
Even though FP attack success rates are still relatively poor, style transfer has at least opened up the possibility of making traditional attack methods amendable. We argue that it is because style transfer can help maintain a better overall scale of the image.

We also rule out the negative effect of baseline methods on the background.
The experimental results indicate that limiting perturbations to the foreground can effectively enhance the performance of the baseline method in TN attacks. However, when it comes to FP attacks, the performance of the baseline method does not show any improvement. This is consistent with our hypothesis that local features can be easily manipulated, while maintaining the consistency of the global scale proves to be challenging. This underlines the difficulty in attacking such models.

In Figure~\ref{f:visRes}, we visually demonstrate the efficacy of different attack methods employed on each dataset.
Illustrations and experimental results show that only our method has successfully executed FP attacks, that common attack techniques fail to accomplish.

In summary, our experimental findings highlight the limitations of traditional adversarial attack methods when applied to signature verification models with clean backgrounds.
Our approach, which emphasizes preserving the overall integrity of the image while making precise modifications, not only offers a more effective means of uncovering potential vulnerabilities but also demonstrates resilience against adversarial training defenses.
This research underscores the importance of developing more nuanced methods to identify and address potential security gaps in this domain.

\subsection{Ablation Studies}

To evaluate the impact of each hyperparameter, we carry out a series of ablation studies.
in terms of setting initial hyperparameters, we aimed to keep the average loss of four functions in an attack within the same range (1e0--1e1).
We tested 50 random samples for each dataset.
For CEDAR, weights were $\alpha$=1e1, $\beta$=1e-3, $\gamma$=1e11, $\delta$=1e2; for BHSig260-B, they were $\alpha$=3e1, $\beta$=1e-2, $\gamma$=2e11, $\delta$=3e1.
Our ablation experiments focused on two metrics: average perturbation and attack success rate.

\begin{figure}[t]
    \centering
    \includegraphics[width=\columnwidth]{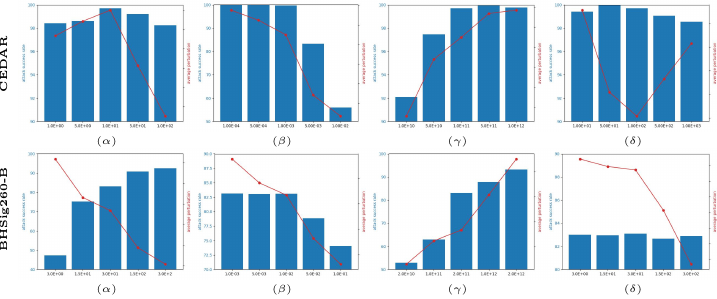}
    \caption{Summary of hyperparameter ablation study results.}
    \label{fig:ablation}
\end{figure}

As shown in Figure~\ref{fig:ablation}, we found that $\alpha$ positively correlates with attack success rate and helps reduce perturbation within a certain range. Increasing $\beta$ effectively lowers perturbation but can also impact the success rate. Experiments with $\gamma$ demonstrated that style transfer consistently enhances attack effectiveness. $\delta$ showed minimal impact on performance but aided in perturbation reduction.

\section{CONCLUSION, LIMITATIONS, AND FUTURE WORK}
In this work, we have introduced two novel loss functions and developed a pioneering method for attacking signature verification models that employ contrastive loss.
By incorporating style transfer principles and our unique loss functions, our approach has outperformed traditional adversarial attack methods, particularly in scenarios with clean backgrounds where conventional methods often falter.

Our research focuses on a model widely recognized in the signature verification community, chosen for its emphasis on similarity matching over traditional classification, aligning with our research goals. We investigate its vulnerability to certain attacks, particularly affecting TN performance, revealing broader risks in machine learning applications. This emphasizes the need for robustness evaluation in specialized models. 

Our method does have some limitations. 
It assumes that attackers have access to a clear, high-quality example of the target signature, which may not always be the case in real-world scenarios.
Additionally, while our technique has been effective against models trained using contrastive loss, it has not been generalized against models based on other loss functions.

Future research should explore ways to overcome these limitations, such as 
developing methods that can work effectively even with lower-quality signature examples or
expanding the approach against models using different loss functions. By addressing these challenges, we can continue to advance the field and develop more secure biometric verification systems.

\section*{Acknowledgments}
We extend our sincere gratitude to the reviewers for their insights on the supplement and design of baselines, which greatly enriched this work.
The first author Zhongliang Guo acknowledges the financial support through the China Scholarship Council – University of St Andrews Scholarship (Grant No.202208060113).
\bibliography{reference}

\end{document}